# Pointer over Attention: An Improved Bangla Text Summarization Approach Using Hybrid Pointer Generator Network

[1]Nobel Dhar, [1]Gaurob Saha, [2]Prithwiraj Bhattacharjee, [2]Avi Mallick, [2]Md Saiful Islam
[1]Department of CSE, Sylhet Engineering College, Sylhet, Bangladesh
[2]Department of CSE, Shahjalal University of Science & Technology, Sylhet, Bangladesh
Email: {nobeldhar807, onabilsahagaurob, avimallick999}@gmail.com, prithwiraj12@student.sust.edu, saiful-cse@sust.edu

*Abstract*—Despite the success of the neural sequence-to-sequence model for abstractive text summarization, it has a few shortcomings, such as repeating inaccurate factual details and tending to repeat themselves. We propose a hybrid pointer generator network to solve the shortcomings of reproducing factual details inadequately and phrase repetition. We augment the attention-based sequence-to-sequence using a hybrid pointer generator network that can generate Out-of-Vocabulary words and enhance accuracy in reproducing authentic details and a coverage mechanism that discourages repetition. It produces a reasonable-sized output text that preserves the conceptual integrity and factual information of the input article. For evaluation, we primarily employed "BANSData"[1] - a highly adopted publicly available Bengali dataset. Additionally, we prepared a large-scale dataset called "BANS-133" which consists of 133k Bangla news articles associated with human-generated summaries. Experimenting with the proposed model, we achieved ROUGE-1 and ROUGE-2 scores of 0.66, 0.41 for the BANSData" dataset and 0.67, 0.42 for the BANS-133k" dataset, respectively. We demonstrated that the proposed system surpasses previous state-of-the-art Bengali abstractive summarization techniques and its stability on a larger dataset. "BANS-133" datasets and code-base will be publicly available for research.

## I. Introduction

Summarization is the technique of condensing text material into a concise version that contains only the most important facts from the input text. It is getting increasingly popular as individuals seek data presented effectively and succinctly for the sake of time management and considerable efficacy. Classification of summarization approaches extractive techniques and abstractive techniques. An extractive method assembles a summary from the original text (approximately full sentences), whereas an abstractive method can generate novel terms in the generated outputs that were not part of the original input text's constitution. Abstractive summaries are distinguished by sophisticated qualities such as real-world senses, rephrasing, and generalization, mainly the prominent features of human-generated summaries. However, abstractive approaches are rarely used due to their technical complications. There are reasonable numbers of high-grade works available based on English abstractive text summarization; very few notable summarization works can be noticed in the Bengali text summarization sector. Furthermore, only a few approaches employ Deep Learning (DL), implying a paucity of appropriate Bengali text datasets for Deep Learning models. So, we decided to prepare a large-scale dataset ("BANS-133") for improved training of our model and create future research scope in this domain. "BANS-133" is the largest Bengali text summarization dataset to date. There are more than 133k articles that we have collected from an online news platform bangla.bdnews24.com[2]. The articles, along with their corresponding summaries, comprise the dataset. Recent abstractive summarization models have centred on headline-generating operations (condensing single or multiple lines of text to a single mainline); it is considered that long-document abstractive summarization techniques are more complex and cumbersome (requiring sublime abstraction quality and discouraging repetition), but finally more efficient and sophisticated. Certain abstractive text summarization systems based on recurrent neural networks (RNNs) unfold good results analyzing and spontaneously generating text. These strategies are promising, but they have flaws, such as the creation of incorrect factual information, incompetence of generating out-of-vocab words, and the limitations of repetitions. For detecting and attempting to resolve these complications in the context of multiple-sentence long document summarization scenarios, we present an enhancement on the sequence-to-sequence attentional model that facilitates the reproduction of flawless information, the effective management of out-of-vocabulary words while maintaining contextual meaning efficiency. We use the coverage method that constantly monitors and analyzes what is being summarized, and it performs well enough by drastically reducing repetition. A brief overview of our contributions:

- Enhanced sequence-to-sequence attention model using a hybrid pointer-generator network with a coverage mechanism that discourages repetition.
- Our proposed system outperformed existing state-of-the-art techniques in terms of quantitative and qualitative evaluation.

---

[1]https://rb.gy/9swova

[2]https://bangla.bdnews24.com/

## II. Literature Review

A large number of summarization algorithms based on a variety of qualitative characteristics have been developed during the last few years. We noticed that, Saggion et al. [1] presented an overview of different summarization techniques along with their evaluation systems. Zazic et al. [2] presented a headline generation approach using hidden Markov models. Then we focused on various corresponding works that were aligned with our subject of study and found out that, Uddin et al. [3] discussed various Bengali text summarization techniques and their implementations. Das et al. [4] advocated the use of theme identification feature in the opinion summarization techniques. Then we noticed Sarkar et al. [5] proposed a singular document extractive summarization approach by using key-phrases features from input text, then this very technique was enhanced by Haque et al. [6] by modifying the key-phrase selection process. Ghosh et al. [7] proposed a Bengali news summarization technique using a graph-based sentence rating method. Sarkar et al. [8] implemented SSS (semantic sentence similarity) and term frequency in their proposed summarizing approach. Haque et al. [9] proposed sentence clustering and sentence frequency for news summarization. Tumpa et al. [10] proposed summary synchronization and also presented the enhanced word scoring process and position value heuristics. Haque et al. [11] proposed an extractive approach using Bengali grammatical regulations, cosine similarity, sentence relevancy and sentence ranking. According to our best knowledge there is only one neural network based Bengali extractive summarization technique, Al Munzir et al. [12] proposed recurrent neural network (RNN) based single document extractive summarization approach. Hereafter we focused on abstractive summarization, Rush et al. [13] the first to use the attention mechanism network based abstractive summarization, attaining remarkable achievement in different metrics on two datasets named Gigaword and DUC-2004, Chopra et al. [14] in thier work, embedded recurrent decoders to that same attention mechanism system to ameliorate the performance on those datasets. In terms of the ROUGE assessment, Nallapati et al. [15] developed a hierarchical RNN-based extractive method that outperformed the previously mentioned abstractive methods. Abujar et al. [16] in their approach, employed word embedding with the word2vec technique. For word embedding Skip-gram and CBOW (Continuous bag of words) models were used. Prithwiraj et al. [17] proposed neural attention based model for Bengali abstractive news summarization. Hasan et al. [18] recently presented abstractive summarization dataset on a large scale for different languages. The main concept of our model "pointer generator network" was proposed by Oriol et al. [19] and we found out See et al. [20] used "pointer generator network" for text summarization. The variant of "Attention mechanism" was adopted from Tu et al. [21] and that performed remarkably in terms of minimizing repetition.

## III. Dataset

A standardized set of data is a crucial component, when it comes to summarizing text. To achieve a better outcome, it is imperative to provide the model with high-quality data. In our search for such quality, we came across the concept Hermann et al. [22] of developing a standard dataset. We prepared "BANS-133" dataset that contained more than 133k articles and their human-generated summaries. Another research work on this dataset is ongoing and after this the dataset will be publicly available for all the researchers.

### A. Dataset Preprocessing

We required a massive volumes of data to train our model, however no single standard big scale dataset for Bengali text summarization has yet been made public. So, we attempted to gather news articles and their corresponding summaries from the internet news platform bangla.bdnews24.com[2]. We were able to collect more than 133k news articles and their summaries from a range of themes using a crawler, including social issues, economic issues, sports, health, lifestyle, and so on. Ads, non-Bengali language, links to other websites, etc. are all common in online newspapers. A data preprocessing system was created to remove all trash from the dataset as a first step in pre-processing the dataset. Then we took other data pre-processing initiatives to make the dataset compatible with our experiment.

### B. Dataset Comparison

A limited number of Bengali text summarization datasets (small scale) have been made publicly available in the previous few years. When we were looking for a good dataset, we came across three of these: In the BNLPC[3] dataset, there are 200 articles, three summaries per article, and 600 total summaries. BANSData[1] dataset, with 19096 total articles,

| Source | Total Articles | Total Summaries |
|---|---|---|
| BNLPC Dataset | 200 | 600 |
| BANSData | 19096 | 19096 |
| {XL}-Sum Dataset | 10126 | 10126 |
| BANS-133 | 133148 | 133148 |

Table I: Comparison of "BANS-133" with existing datasets

1 summary per article, and 19096 total summaries whereas {XL}-Sum[4] has 10126 articles, 1 summary per article, and 10126 total summaries. Our dataset "BANS 133" has 133148 articles, with one human created summary per article and a total of 133148 summaries, making it a substantial one when compared to the existing other Bengali text summarization datasets. Table I exhibits the comparison among "BANS-133" and other existing datasets.

---

[3]http://www.bnlpc.org/research.php
[4]https://github.com/csebuetnlp/xl-sum

## IV. Model Architecture

Recurrent neural networks (RNNs) have become ideal for many natural language processing tasks that deal with sequential data over the recent few years. Particularly, the sequence-to-sequence model with attention has been very well appreciated for summarization. Nevertheless, this approach encounters two big obstacles, reproducing inaccurate factual details for rare or out-of-vocabulary words and repetition of words. Our model solves the inaccurate copying problem by using the pointer-generator network proposed in [20]. Via pointing, this hybrid network can choose to replicate words from the source while hanging onto the ability to generate words from the fixed vocabulary. The pointer generator network's architecture is depicted in the diagram below:

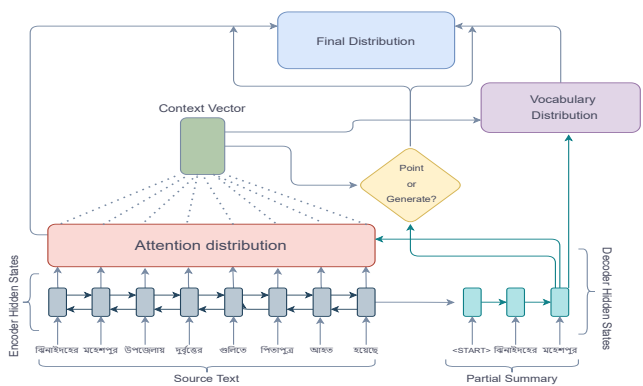

Figure 1: Model Architecture [20, p. 3]

### A. Encoder & Decoder

The data point we used in the example begins with "ঝিনাই-দহের মহেশপুর উপজেলায় দুর্বৃত্তের গুলিতে পিতাপুত্র আহত হয়েছে", and we are in the process of producing the summary "ঝিনাইদহের মহেশপুরে দুর্বৃত্তের গুলিতে পিতাপুত্র আহত". At first, the source text is fed word-by-word to the encoder RNN, which generates a series of encoder hidden states. After the encoder RNN has finished reading the original input text, the decoder RNN generates a series of words to construct a summary. The decoder collects the previous word of the summary as its input (at the first phase, this is a unique <START> symbol, that is the indication to starting of the sequence) and employs it to update the hidden state of the decoder. The attention distribution over the original input words is computed using this information.

### B. Attention distribution

Attention distribution is a probability distribution that helps the network to decide where to look in the input sequence to generate the next word. This model uses the attention technique to determine the relevance of the currently processing word and the next words in the input sequence, then generates the word with the highest probability distribution. According to the example data point of figure 1: the decoder had generated the very first word 'ঝিনাইদহের' and then focused on rest of the words 'মহেশপুর', 'উপজেলায়', 'দুর্বৃত্তের', 'গুলিতে', 'পিতাপুত্র', 'আহত', 'হয়েছে' of input sequence and abstractly generated the word 'মহেশপুরে' using the attention distribution (a word with highest attention distribution gets generated early). It is now focusing on the rest of input sequence words 'উপজেলায়', 'দুর্বৃত্তের', 'গুলিতে', 'পিতাপুত্র', 'আহত', 'হয়েছে' and will generate the word 'গুলিতে' with highest attention distribution value. This is how attention-distribution is used.

### C. Context Vector

The context vector is created using attention distribution, a weighted sum of the encoder hidden states. It records what was read from the original text. Following that, the vocabulary distribution is determined using the context vector and decoder hidden state.

### D. Pointer-Generator

In section IV(B) and IV(C), we calculated attention distribution and vocabulary distribution. In this section, we calculate generation probability, which allows us to combine the attention and vocabulary distribution into a final distribution. Generation probability stands for the likelihood of generating a word from the vocabulary versus copying a word from the original text [20]. The value of generation probability rests between 0 and 1 [20]. Later on, based on the final distribution, the model decides whether it needs to generate a word or point to it anywhere it is available in the source text. This is where it surpasses all the other state of the art models. Until this, a model could either point to a word from the source or generate a word from the vocabulary but could not utilise both techniques according to the need. Moreover, the pointer-generator system simplifies replicating words from the original text using pointing, which increases precision and manages out-of-vocab terms while managing the capability to produce novel words.

### E. Coverage Mechanism

To tackle repetition of words, a technique called "coverage" is used. The idea is that, the attention distribution is used to keep track of what's been covered so far, and penalize the network for attending to same parts again. According to [21] to construct the coverage mechanism, each input word's annotation is attached to a coverage vector. The value of each vector is initially set to zero, but it is modified after each "attentive-reading" (i.e. during decoding, the process of giving attention to most probable words to be generated) of the corresponding word annotation. This vector is the input into the attention mechanism to aid in the preference of subsequent attention, allowing the system to consider more non-generated input words. If any word receives more attention, the coverage vector will discourage increased attention to it and as a result, will guide attention to the less attended portions of the input sentence. The coverage mechanism uses this strategy to reduce duplication.

| Actual Article | Reference Summary | Attention Summary (without Coverage) (BANS) [17] | Pointer Generator Model + Coverage Summary |
|---|---|---|---|
| অবশেষে ইউরোপে প্রাই-ভেসি পলিসি পরিবর্তনে রাজি হয়েছে ওয়েব জা-য়ান্ট গুগল। ব্যবহারকারী-দের ডেটা কীভাবে সংগ্রহ করা হচ্ছে সে বিষয়টি ব্য-বহারকারীদের কাছে আরও সহজ করে বলতে হবে প্র-তিষ্ঠানটিকে। | তথ্য পাওয়া সহজ করতে গুগলের পদক্ষেপে | গুগল আরও আরও সহজ | গুগল আরও সহজ |
| রাষ্ট্রপতি মো আবদুল হামি-দের সঙ্গে সৌজন্য সাক্ষাৎ করেছেন প্রধান বিচারপতি এস কে সিনহা। | রাষ্ট্রপতির সঙ্গে প্রধান বিচার-পতির সাক্ষাৎ | রাষ্ট্রপতির সঙ্গে সঙ্গে সাক্ষাৎ করেছেন প্রধান বিচারপতি সিনহা | রাষ্ট্রপতির সঙ্গে সাক্ষাৎ করে-ছেন প্রধান বি-চারপতি সিনহা |
| মুহিত বলেছেন,বিশ্ব ব্যাং-কের কাছ থেকে সহজ শর্তে কম সুদে ঋণ নে-ওয়ার সুযোগ থেকে ধীরে ধীরে বেরিয়ে আসতে চায় বাংলাদেশ। | একটু বেশি সু-দে হলেও বড় ঋণ চাই : মু-হিত | বিশ্ব ব্যাংকের সহজ ঋণ সুযোগ চায় | বিশ্ব ব্যাংকের সহজ ঋণ সু-যোগ চায় বাং-লাদেশ |

Table II: Illustrates some of the summaries of attention model (BANS) [17] and pointer-generator model.

## V. Result Analysis

For assessing the proposed system, we evaluated our model using two variations of rating matrices: quantitative and qualitative. Both the assessment approach was important to evaluate the model capable of producing a good summary. Following the standard convention or general rule of thumb, seventy percent of the data was applied for model training, twenty percent for model validating, and ten percent for model testing. Five times, we trained our system, every time with a different set of parameters. The system produced the best results when the vocab size was fixed at 50k, hidden states dimensions at 512, embedding dimensions at 128, maximum encoding steps at 400, maximum decoding steps at 100, learning rate at 0.15. Table II demonstrates some of the cases that our model (Pointer Generator + Coverage) has produced alongside the corresponding attention cases. For the first summary, our model (Pointer Generator + Coverage) shows ambiguity by producing a generalized output "গুগল আরও সহজ", whereas the main article is strictly associated with the specific term "ইউরোপে". The first attention (BANS) [17] summary additionally exhibits repetition "আরও আরও" alongside with this problem. The second one "রাষ্ট্রপতির সঙ্গে সাক্ষাৎ করেছেন প্রধান বিচারপতি সিনহা" generated by our model is a flawless summary. Whereas attention model (BANS) [17] generates factual error "রাষ্ট্রপতি সঙ্গে" and repetition "সঙ্গে সঙ্গে". The third one of our model "বিশ্ব ব্যাংকের সহজ ঋণ সুযোগ চায় বাংলাদেশ" demonstrates sheer abstractness along with preserving contextual meaning. The model can discern the implication of the term "সহজ শর্তে কম সুদে ঋণ" and generates the complete explicit form "সহজ ঋণ সুযোগ চায়", which is not stated in the main article. The third summary "বিশ্ব ব্যাংকের সহজ ঋণ সুযোগ চায়" of attention model (BANS) [17] is incomplete, missing the subject of the sentence. These examples justify the supremacy of the Pointer-Generator over the Attention model.

### A. Quantitative Evaluation

A system-oriented assessment is quantitative evaluation. In this study, both the actual and anticipated summaries are fed into an algorithm, which assigns a score depending on how much the anticipated summary differs from the actual summary. We discovered two common quantitative assessment matrices: ROUGE and BLEU. Since ROUGE scoring is a popular and standard approach for measuring summarization quality, we opted for this evaluation process for our experiment. The ROUGE-1, ROUGE-2, ROUGE-L score was calculated. We started by calculating the Precision and Recall for the ROUGE method. The average F1 score for 100 occurrences was calculated using these two metrics. Our final score was significant according to the article's [23] standard evaluation and surpassed publicly available {XL}-Sum [18] and BANS 19k [17] ROUGE scores. Figure 2 illustrates the comparison between ROUGE scores of {XL}-Sum [18], BANS 19k [17], pointer 19k and pointer 133k.

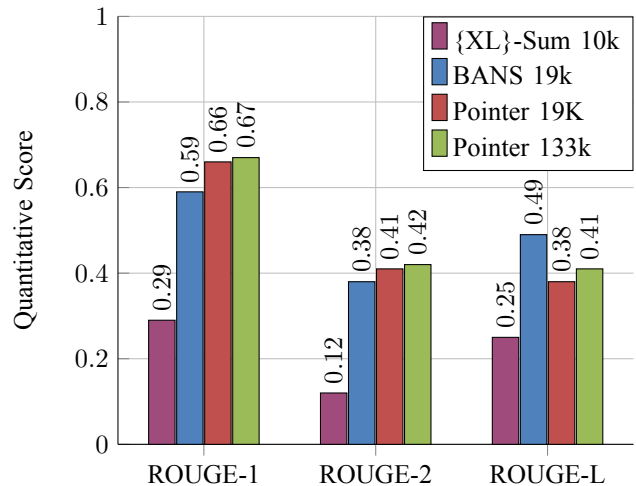

Figure 2: Illustrates the comparison of ROUGE-1, ROUGE-2 and ROUGE-L Scores among different existing models and proposed pointer models

### B. Qualitative Evaluation

The user-centred evaluation approach is known as qualitative evaluation. Some users of various ages were asked to assess the generated summary on a scale of 1 to 5 compared to the original summary. We collected the qualitative evaluation scores of BANS [17] and Bi-LSTM [24] from BANS [17]. Then using a google form[5] survey, reviewers were shown some of the output samples of our proposed systems. The corresponding input samples were taken from both "BANSData" and "BANS-133" datasets. A total of 20 people used the 5-point rating system. Additionally, all of the users were between the ages of 23 and 27 and originated from various educational histories. Then the average rating was determined. We compared the average rating to the publicly available scores of each existing model and found that our method outscored them in human assessment. Table III displays publicly released human evaluation points for state-of-the-art models as well as our model's attained human evaluation points.

---
[5]https://forms.gle/hN1xhqBQHnnSubit5

| Model | Average Rating (Out of 5) |
|---|---|
| Bi-LSTM [24] | 2.75 |
| BANS [17] | 2.80 |
| Pointer on BANSData | 3.13 |
| Pointer on BANS-133 | 3.18 |

Table III: Qualitative Evaluation of existing models and the proposed pointer-generator model

## VI. Conclusion

Our most important contribution is improving the sequence-to-sequence attentional model employing a hybrid pointer-generator network with a coverage mechanism that discourages repetition. Then preparation of the "BANS-133" dataset, which contains over 133k articles and their human-generated summaries. Qualitative ratings indicate a significant improvement in summarization quality. On two larger-scale datasets ("BANSData" and "BANS-133"), our model's ROUGE scores excel existing state-of-the-art Bengali text summarizing architectures. Even though our technology generates abstract summaries, the language is frequently equivalent to the original document. Occasionally, the network fails to emphasize the crux of the original text, instead opting to summarize a less relevant, auxiliary piece of information. However, despite our best efforts, these are some difficulties that we hope to solve in future initiatives.

## VII. Acknowledgement

We want to thank all the volunteers who helped us to collect and prepare the dataset. We want to show our gratitude to the Shahjalal University of Science and Technology (SUST) research centre and the SUST NLP research group for their support.